\documentclass{bmvc2k}
\usepackage{xurl} 
\usepackage{algorithm,algpseudocode}
\title{Rotating-star Pattern for Camera Calibration}

\addauthor{Zezhun Shi}{spdfghizzs@gmail.com}{1}

\addinstitution{
 Individual Researcher
}

\runninghead{Shi}{Rotating-star Pattern for Camera Calibration}

\def\etal{\emph{et al}\bmvaOneDot}
\begin{document}

\maketitle

\begin{abstract}
Camera calibration is fundamental to 3D vision, and the choice of calibration pattern greatly affects the accuracy. To address aberration issue, star-shaped pattern has been proposed as alternatives to traditional checkerboard. However, such pattern suffers from aliasing artifacts. In this paper, we present a novel solution by employing a series of checkerboard patterns rotated around a central point instead of a single star-shaped pattern. We further propose a complete feature extraction algorithm tailored for this design. Experimental results demonstrate that our approach offers improved accuracy over the conventional star-shaped pattern and achieves high stability across varying exposure levels.

\end{abstract}

%-------------------------------------------------------------------------
\section{Introduction}

\begin{figure*}[h!]
	\begin{center}
		%\fbox{\rule{0pt}{2in} \rule{.9\linewidth}{0pt}}
		\includegraphics[width=0.8\linewidth]{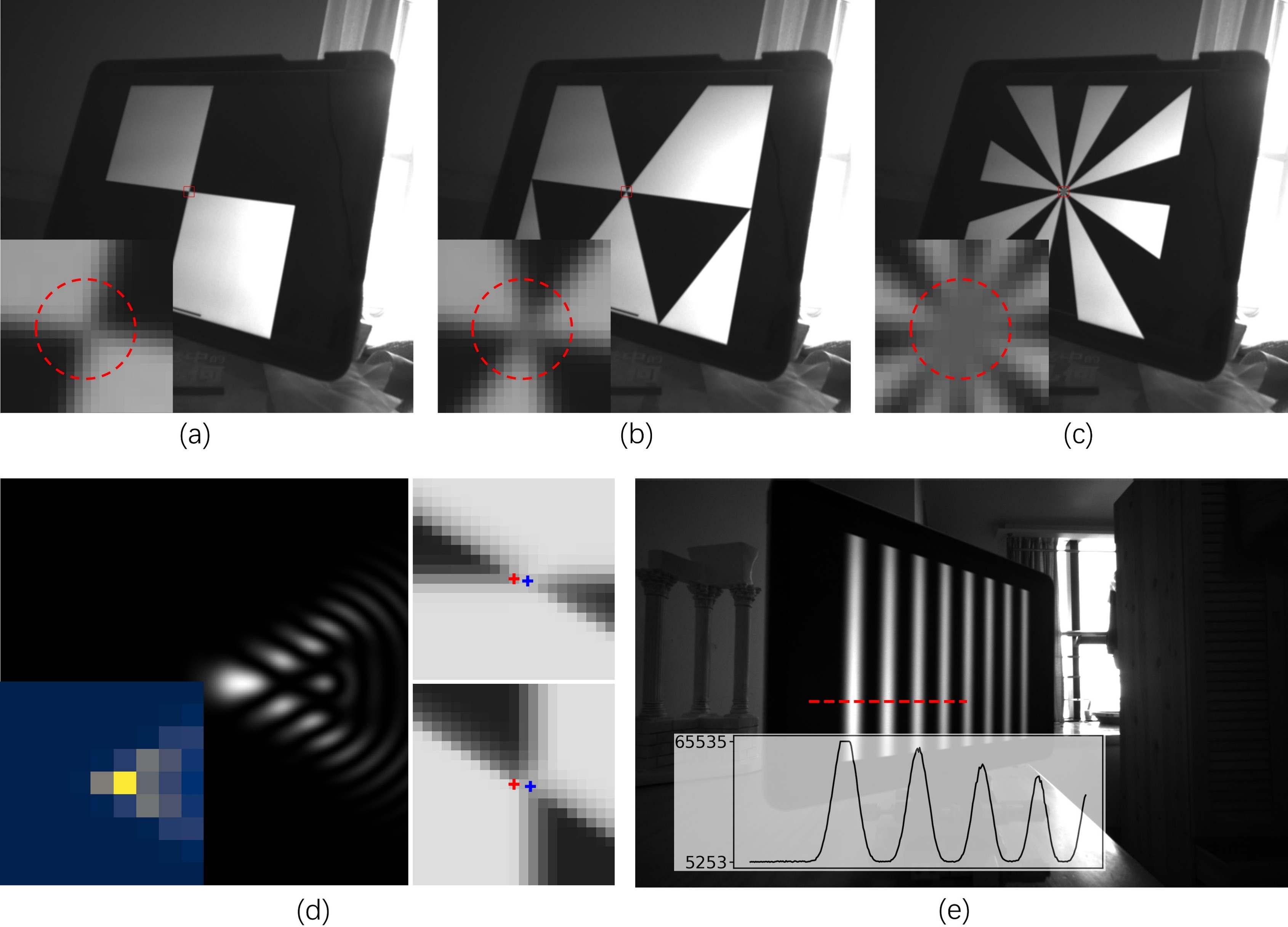}
	\end{center}
	\caption{
		(a) to (c): checkerboard, deltille~\cite{Ha2017}, and star pattern~\cite{Schops2020}.
		(d): Simulation of checkerboard corners under primary coma aberration~\cite{Wyant1992,Wyant2013}. Red markers: aberration free center. Blue markers: OpenCV sub-pixel results. (e): Saturated phase pattern.}
	\label{fig:motivation}
\end{figure*}

\label{sec:intro}
Camera calibration is the foundation of visual applications, determining the precision limits of almost all 3D vision algorithms, such as stereo~\cite{Fusiello2000,Hirschmuller2005,Barnes2009} and structure from motion~\cite{Hartley2003,Schonberger2016}. Camera calibration involves establishing the correspondence between pixels and light rays. The mainstream approach is to extract feature points from an artificial calibration target and then map these to their actual physical positions to fit a camera model~\cite{Brown1971,Brown1986,Zhang2000,Grossberg2001,Kannala2004,Scaramuzza2006,Dunne2007,Ramalingam2016,Schops2020}. Its accuracy mainly comprises three aspects: the calibration pattern, the fitting capability of the camera model, and the precision of feature extraction.

Generally speaking, a calibration pattern with higher physical precision is always preferable, but the choice of calibration pattern can also significantly affect the results. Commonly used patterns include circles~\cite{Heikkila2000,Datta2009,Circles1}, checkerboards~\cite{Lucchese2002,Geiger2012,Placht2014,Duda2018,CurvedCheckboard} and others~\cite{Ha2015,Ha2017,Schops2020,Phase1,Phase2,Phase3}. Checkerboard detection uses local corner points for localization, is better suited for handling high distortion than circles; however, due to the smaller amount of image information utilized, its detection accuracy tends to be lower, and more easily affected by anisotropic aberrations(Fig.~\ref{fig:motivation}(d)). Existing improvement methods include more robust sub-pixel algorithms~\cite{Placht2014,Duda2018} and the use of patterns with additional intersecting lines such as deltille pattern~\cite{Ha2017} and star pattern~\cite{Schops2020}(Fig.~\ref{fig:motivation}(b-c)).

Although the star pattern intuitively provides more gradient information and better handles aberrations, increasing the number of intersecting line patterns introduces more aliasing near the features(Fig.~\ref{fig:motivation}(c)), which necessitates a larger feature optimization window, thereby largely negating the advantage of the checkerboard pattern in handling distortion.

Motivated by this observation, we propose a novel active calibration pattern that decomposes a single star-shaped pattern into a sequence of star patterns with reduced line intersections. And we alternate the pattern boundaries between black and white to enable more robust initialization. In synthetic experiments, while prior works typically simulate lens blur using simple Gaussian functions, we adopt physically realistic primary aberrations, revealing how the symmetry of the point spread function (PSF) influences the accuracy of feature refinement. To assess real-world applicability, we further evaluate the proposed method under varying exposure and calibration board poses, where it shows superior accuracy and robustness over baseline methods.

\section{Related works}
\label{sec:related_works}
\noindent \textbf{Detection of checkerboard pattern and its generation.} F\"orstner and G\"ulch~\cite{Forstner1987} give a corner sub-pixel algorithm, which base on assumption that the gradient direction of each pixel should be orthogonal to the vector pointing from the pixel to the corner in ideal case. This algorithm was adopted by OpenCV. However, such gradient-based approaches are often sensitive to noise in practical scenarios. To address this, Duda and Frese~\cite{Duda2018} proposed a detection method based on the Radon transform and Gaussian peak fit, which shows superior robustness compared to OpenCV in challenging outdoor environments.

Another class of methods that do not rely on image gradients was introduced by Lucchese and Mitra~\cite{Lucchese2002}, who modeled the corner region as a saddle surface. This idea was adopted in the MATLAB toolbox, which incorporates a preprocessing step from Geiger~\etal~\cite{Geiger2012}. Other preprocessing scheme including method proposed by Placht~\etal~\cite{Placht2014} with a cone filter. The saddle surface fitting method was also extended by Wang~\etal~\cite{CurvedCheckboard} to handle checkerboard patterns embedded on curved surfaces.

Ha~\etal~\cite{Ha2017} further generalized saddle fitting method to the case of monkey saddle corners, where three lines intersect(Fig.~\ref{fig:motivation}(b)). Building on Ha’s pattern design, Sch\"ops~\etal~\cite{Schops2020} introduced a multi-line intersection pattern(Fig.~\ref{fig:motivation}(c)) to improve robustness by consideration of chromatic aberration. Their experimental results showed that introducing more intersecting lines significantly reduced reprojection error within a certain range. However, due to pattern aliasing, the maximum number of black-white segments is limited to approximately 16–20, exceeding this range results in a gradual increase in reprojection error.

\noindent \textbf{Detection of active pattern.}
Structured light achieves high-precision 3D reconstruction by projecting specific patterns onto objects~\cite{Salvi2004}. In recent years, some researchers have utilized structured light patterns displayed on screens for camera calibration. Ha~\etal~\cite{Ha2015} proposed a method where a one-dimensional complementary black-and-white stripe pattern is displayed on a mobile phone screen, followed by Gaussian point spread function deconvolution and screen refraction correction, enabling accurate calibration under extreme close-range defocus conditions. However, the assumption of a Gaussian PSF limits the applicability of this approach~\cite{Joshi2008,Brauers2010,FisheyePSF2018}. 

Some researchers have adopted the classical phase-shifting method, using multi-frame pixel values to compute phase as feature points~\cite{Dunne2007,Phase1,Phase2,Phase3}. Many of these studies have demonstrated, through Gaussian blur simulations, that the phase-shifting method is more robust to defocus than checkerboard pattern, and real-world experiments have also shown some accuracy advantages. However, the phase-shifting method relies on absolute mathematical relationships of pixel values, making it susceptible to issues such as ambient lighting, noise, aberration, and pattern resolution. Additionally, overexposure is sometimes a issue in practice(Fig.~\ref{fig:motivation}(e)), although this issue can be addressed by using a greater number of patterns~\cite{Phase4}, the huge required number making it difficult to apply in practice.

\section{Methodology}
% 1. 为什么做 2.做了啥 3.为什么这么做，做别的好不好
\label{sec:method}
\subsection{Pattern Design}
As Sch\"ops~\etal~\cite{Schops2020} mentioned, the checkerboard pattern may affect by aberrations, which can also be seen in the related literature on PSF estimation~\cite{Joshi2008,Brauers2010}.  Therefore, they extended the checkerboard pattern to a star-shaped pattern, where the checkerboard can be viewed as a star with the minimum number of intersecting lines. Although star pattern is a good solution for handling aberration and noise, due to the effects of aliasing, the number of intersecting lines in the star-shaped pattern cannot increase indefinitely, and the aliasing issue(Fig.~\ref{fig:motivation}(c)) can make it unsuitable for high-distortion lens calibration. 
\begin{figure*}[h!]
	\begin{center}
		%\fbox{\rule{0pt}{2in} \rule{.9\linewidth}{0pt}}
		\includegraphics[width=0.99\linewidth]{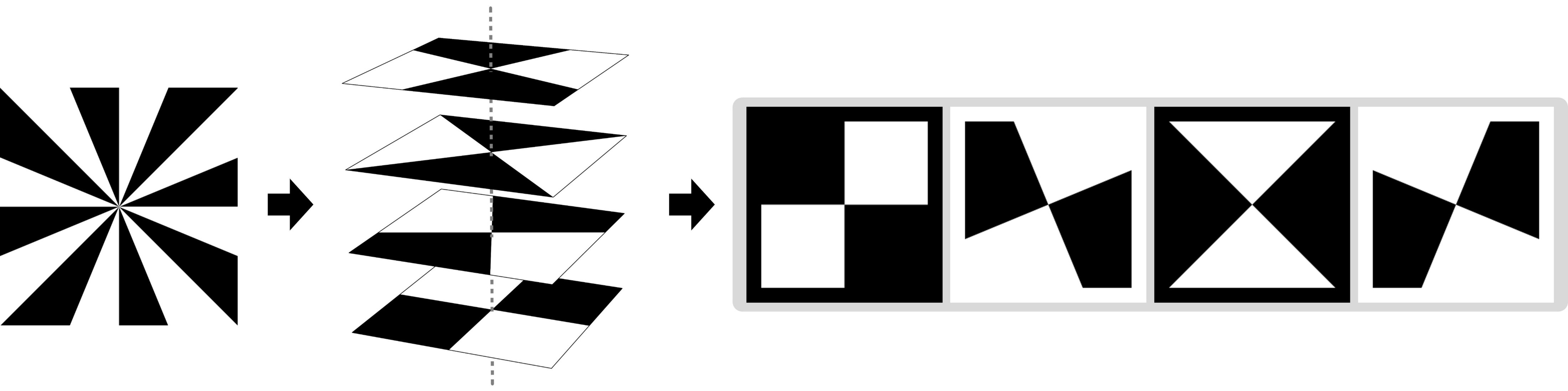}
	\end{center}
	\caption{We decompose the star-shaped pattern into a series of checkerboard patterns rotating around its local center, displayed by a screen. And we add black-and-white transition boundaries around the pattern elements for more robust detection.}
	\label{fig:pattern}
\end{figure*}

Our approach is to replace the single star-shaped pattern with a series of simpler star patterns rotated around the center.  In our experiment, it sets to two intersection lines(see Fig.~\ref{fig:pattern}), i.e., checkerboard pattern. This allows us to perform more robust detection to aberrations and noise than checkerboard, while we can obtain more gradient information within a smaller optimization window than single star-shaped pattern. Further decomposing into black-and-white stripes would double the data acquisition time, and there are already many well-established algorithms for refining the corners of a checkerboard. In experiments, we found that existing checkerboard detection algorithms are quite sensitive to distortion and illumination change, sometimes leading to detection failures. Thus we add alternating black and white boundary lines around the pattern elements in two adjacent frames, which allows us to robustly determine the bound of the pattern elements through accumulating the pixel differences between adjacent frames. The width of bound is set in proportion to minimum optimization window we choose. This allows for better utilization of the redundant pixels that are not involved in the refinement process.

\subsection{Corner Initialization}
The goal of initialization is to obtain a rough estimate of feature point locations and establish their correspondence with world coordinates. In this paper, we focus on the detection of the complete pattern. Thus we simply organize the feature points in a 2D array, same as OpenCV and MATLAB. Our initialization process consists of three steps:
\begin{itemize}
	\item Computing the pattern boundary. We accumulate the absolute differences between adjacent frames for each pixel, binarize the result, and extract contours from the binary image. The longest contour along with its enclosed sub-contours defines boundary of pattern elements.
	\item Obtaining rough locations. For each boundary, we extract edge points within the largest inscribed square centered at the centroid, and fit two quadratic curves to them with RANSAC(group points by fitting lines first). The rough locations are determined by a least-squares optimization over all frames to find the points closest to these quadratic curves.
	\item Establishing world coordinate correspondence. We construct a graph by brute-force enumerating all pairs of elements. For each pair, if the minimum distance between the bounding rectangle's vertices is less than half the length of the rectangle's shortest side, connect the two nodes with an edge. Using neighbor counts and adjacency, boundary nodes are extracted and arranged into a linked list. Once the world coordinates of the four corners are set, those of the other nodes in linked list and adjacent interior nodes can be inferred. All remaining nodes will then be handled recursively.
\end{itemize}

Binarization is a crucial step for correctly obtaining the initial corners. In our experiments, we adopted a very simple approach: due to uneven illumination, we first compute a per-pixel maximum over all frames to obtain an intensity estimation. To avoid divergence caused by division by zero, pixels in this estimation with values lower than a threshold(1/20) of the global maximum are then inverted. Next, we accumulate the differences between adjacent frames and divide the result by the intensity estimation to generate a response map. Finally, mean thresholding is applied for binarization.

\subsection{Corner Refinement}
Sch\"ops~\etal~\cite{Schops2020} give a symmetry based feature refinement approach which shows remarkable result. Inspired by this, we adopt the following optimization objective
\begin{equation}
	\arg\min_{\mathbf{q}}\sum_{\Delta_i}\left[\mathcal{I}(\mathbf{q}+\Delta_i)-\mathcal{I}(\mathbf{q}-\Delta_i))\right]^2,
	\label{eq:cost}
\end{equation}
where $\mathcal{I}$ is image value and $\mathbf{q}$ is feature point. Similar to Sch\"ops~\etal, we randomly sample sub-pixel points $\Delta_i$ and then optimized it by LM algorithm. The Jacobian is calculated by bilinear interpolation of the discrete gradient as well. We choose this cost function because it is faster to compute, theoretically yields good results for isotropic PSFs, and also performs very well for asymmetric PSFs in our  synthetic experiment. 
%A similar approach can also be found in the supplementary materials of Ref.~\cite{Schops2019} where gradient was used.

\section{Experiment}
This section includes both synthetic and real-world experiments.
Lenses with significant aberrations may not be adequately modeled by easy to use parametric camera models. Therefore, we employ synthetic experiment to assess the stability of feature extraction algorithms across different viewing angles at the same physical location, instead of relying on reprojection error. In the real-world experiments, we compare the proposed pattern against the star-shaped pattern and the phase pattern. We primarily focus on testing the calibration results from independent views, with consideration of the exposure settings and the tilt angle of the calibration board.

\subsection{Synthetic experiment}
\begin{figure*}[h]
	\begin{center}
		\includegraphics[width=0.8\linewidth]{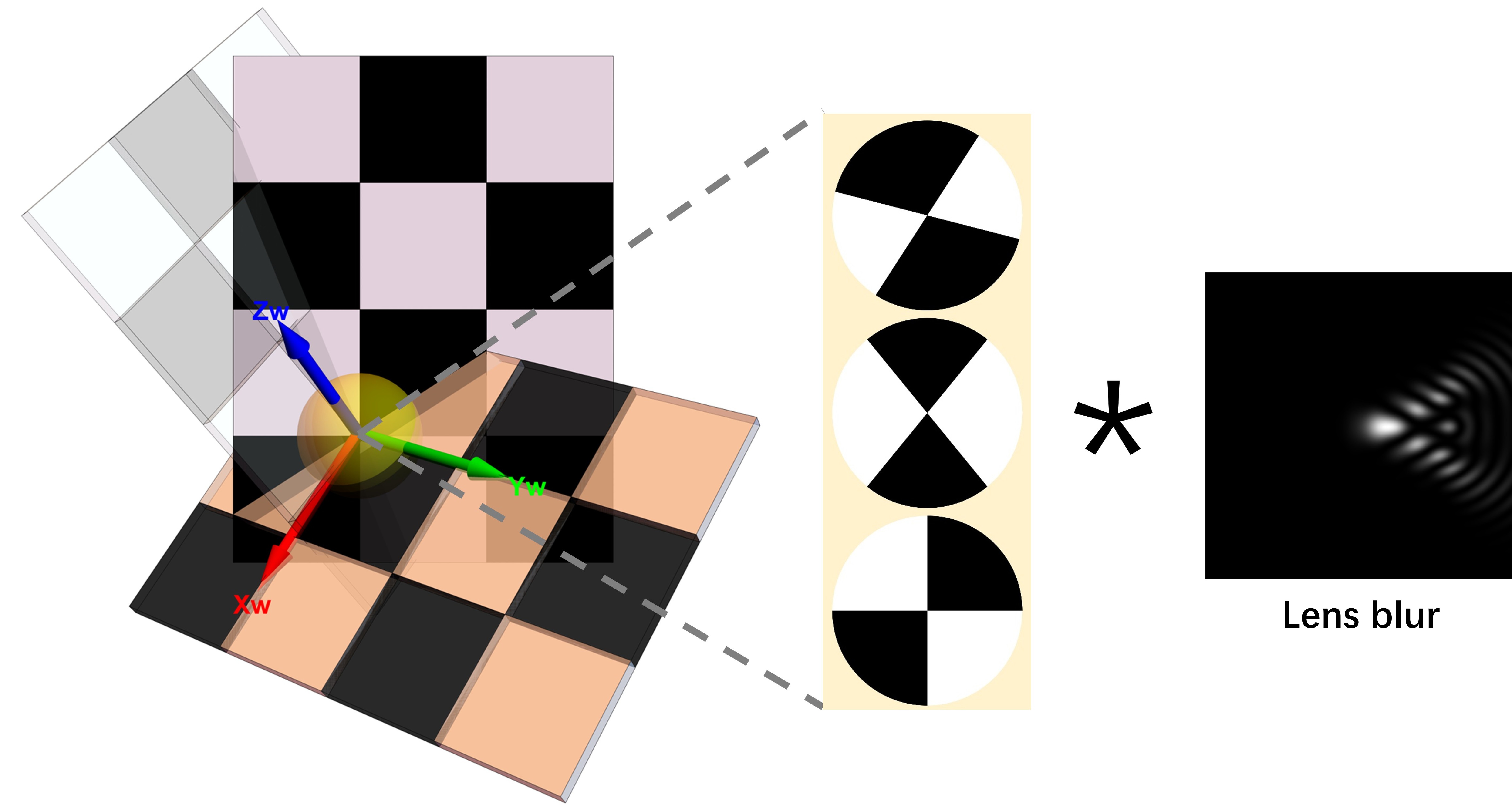}
	\end{center}
	\caption{Model for synthetic experiment.}
	\label{fig:synthetic_model}
\end{figure*}
We bind the world coordinate system to the checkerboard corners, fix the camera viewpoint, and rotate the checkerboard around the origin of the world coordinate system(Fig.~\ref{fig:synthetic_model}). Initially, the Z-axis of the world coordinate system is aligned with that of the camera. For each rotation, the checkerboard is first rotated by an angle $\phi$ around the world Z-axis from initial pose, followed by a rotation by an angle $\alpha$ around the X-axis. The angle $\phi$ varies from 10\textdegree~to 60\textdegree~in 10\textdegree~increments, and 
$\alpha$ ranges from 0\textdegree~to 165\textdegree~in 15\textdegree~increments. The rotating-star pattern undergoes the same rotational motion, and its initial frame is aligned with the checkerboard.

In the simulation, we take aberrations into account. Specifically, the wavefront is a surface representing points of a wave that have the same phase, the first-order wavefront properties and third-order wavefront aberrations can be obtained from the Zernike polynomials and expressed in terms of aberration coefficients as~\cite{Wyant1992,Wyant2013}
\begin{equation}
	\begin{aligned}
		W(\rho,\theta)&=W_{tilt}\rho\cos(\theta)+W_{focus}\rho^2\\
		&+W_{ast.}\rho^4+W_{coma}\rho^3\cos(\theta)+W_{sph.}\rho^2\cos(\theta)^2
	\end{aligned},
	\label{wave_front}
\end{equation}
where $\rho$ and $\theta$ are normalized position within the pupil. This is called Seidel aberration or primary aberration. Each term of above equation is one of aberrations: tilt, focus, astigmatism, coma and spherical. The corresponding PSF can be calculated as 
\begin{equation}
	H(x,y)=A\left|\mathcal{F}\left(e^{2\pi i W(\rho,\theta)}\right)\right|^2
	\label{psf},
\end{equation}
where $\mathcal{F}$ is Fourier transform and $A$ is normalization factor. An example of coma PSF is shown in Fig.~\ref{fig:motivation}(d) with $W_{coma}=4$. 

For each aberration, we calculate the PSF of wavefront by Eqs.~\eqref{wave_front} and~\eqref{psf}, then convolve it with the checkerboard and rotating-star, and finally discretize it by performing block-wise summation. Gaussian noise $\sigma_n$ is then added to the pixel values. As in the following real experiment, we use a rotating-star sequence of 8 images. For a fair assessment of noise effects, we generate 8 noisy versions of the same checkerboard pattern with independent random noise and average them. The evaluation is based on the pairwise-distance range, where we enumerate all pairs of detected points, calculate their distances and select the maximum value. To account for randomness, the process is repeated 50 times, and the average is taken.

\begin{figure*}[h!]
	\begin{center}
		\includegraphics[width=0.99\linewidth]{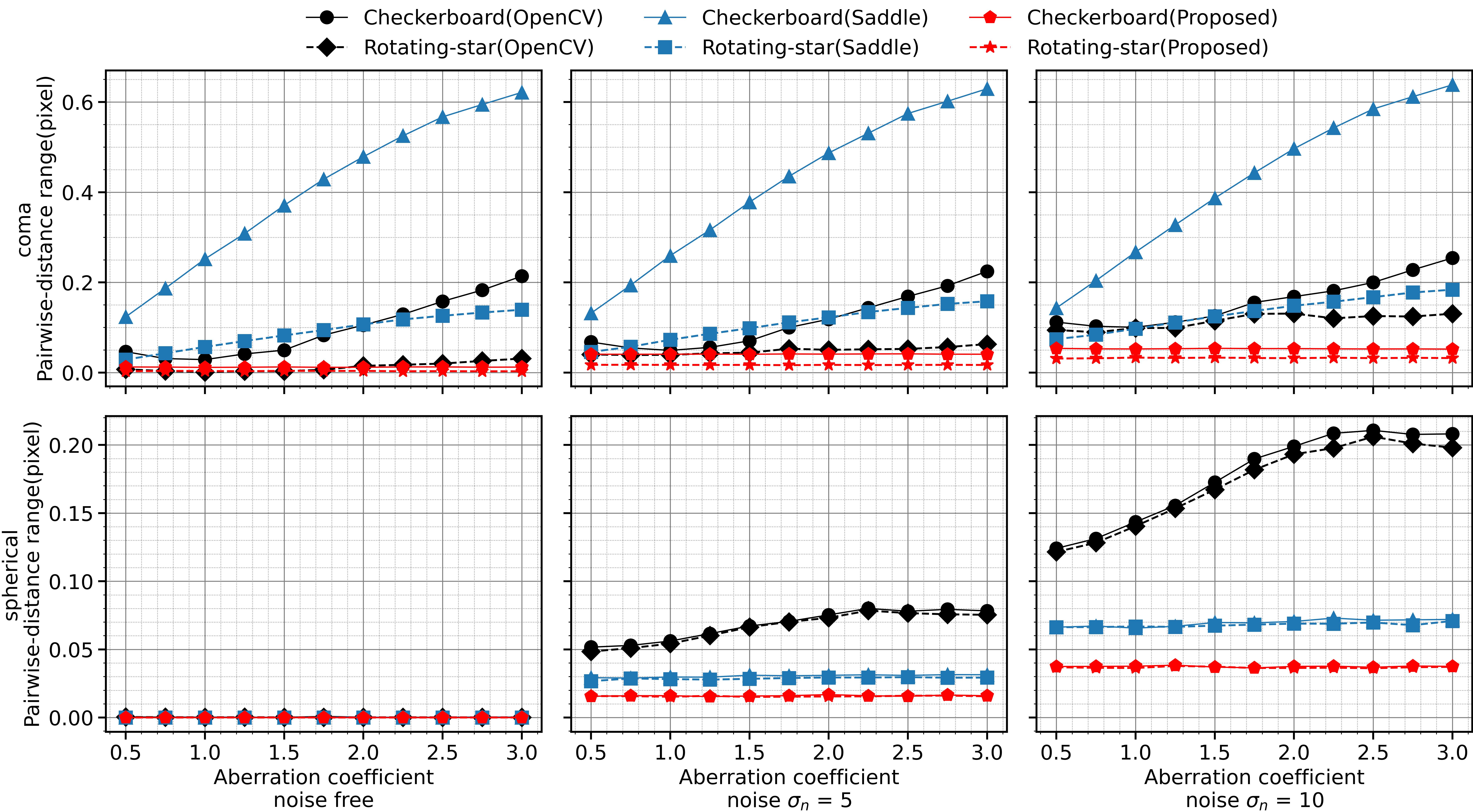}
	\end{center}
	\caption{Pairwise-distance range of refinement algorithms under primary aberration.}
	\label{fig:synthetic_result}
\end{figure*}

In Fig.~\ref{fig:synthetic_result}, we present the results of three algorithms under spherical and coma aberrations. Other types of aberrations, due to their central symmetry, produce results similar to those under spherical aberration. The three algorithms are: the gradient-based method used by OpenCV~\cite{Forstner1987}, the saddle surface fitting method~\cite{Lucchese2002}, and proposed refinement method described in Eq.~\eqref{eq:cost}. Since both the OpenCV and proposed method involve iterative procedures, to avoid the trivial case where no updates occur, we introduce a random perturbation of one pixel to the initial x and y values(ground truth) of these two methods.

The real situation may involve a mixture of various aberrations, including higher-order aberrations, complex illumination, pattern blur... A complete simulation of the optical system is not our goal. However, we can still observe some interesting impacts of PSF symmetry on algorithm performance, which cannot be seen in simple Gaussian assumption experiments: 
\begin{itemize}
	\item In the noise-free case, the asymmetry of the PSF can significantly affect the stability of the corner refinement algorithms.
	\item Under symmetric PSFs, the rotating-star pattern performs almost equivalently to the averaged checkerboard images. In contrast, under asymmetric PSFs, it brings significant improvements for all algorithms.
	\item With a symmetric PSF, the saddle surface fitting method is more robust to noise than the gradient-based method. Otherwise, saddle surface fitting is worse. The proposed refinement algorithm achieves the best results under both types of PSFs.
\end{itemize}

The notably poor results of the saddle surface fitting method under asymmetric PSFs can be largely attributed to the initial corner positions. In our experiment, we simply kept the initial values consistent across all methods, as we think that the robustness to initial values should also be considered as part of the algorithm evaluation.

\subsection{Real world experiment}
\begin{figure*}[h!]
	\begin{center}
		\includegraphics[width=0.99\linewidth]{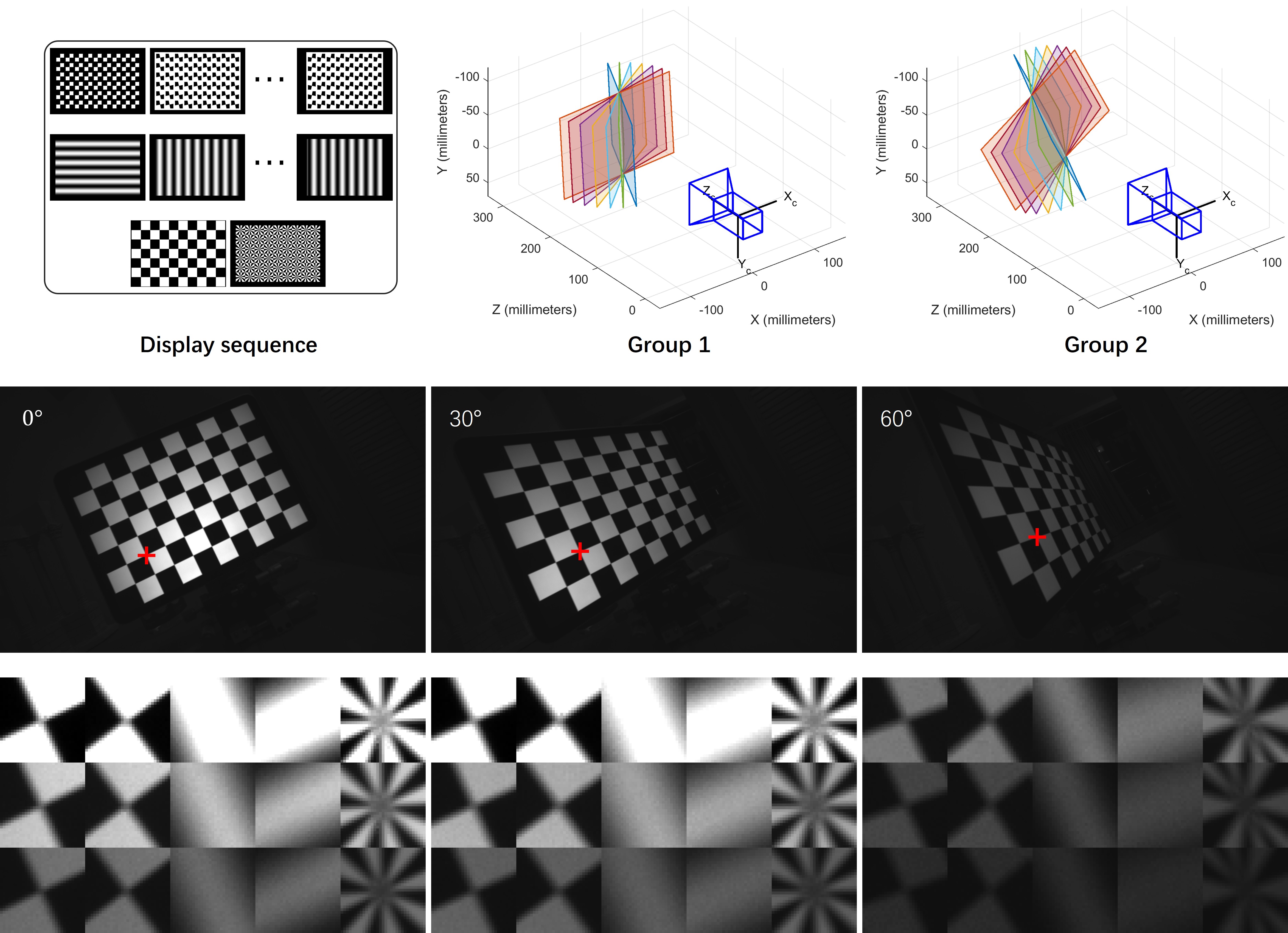}
	\end{center}
	\caption{(Top row) Display sequence and poses of two calibration groups. (Middle row) Captured checkerboard images of the right group under middle exposure level. (Bottom row) Close-up of patterns under three exposure levels at the red cross location of the middle row.}
	\label{fig:ext_normal}
\end{figure*}

In the real world experiment, an iPad with a resolution of $2360\times1640$ was used to play a sequence of patterns. The sequence consists of 8 rotating-star images, each rotated by $\pi/16$ relative to the previous one, followed by 8 images with 4-step phase shifts in the x and y directions, and finally, a checkerboard and star pattern(see top row of Fig.~\ref{fig:ext_normal}). For the same position, the average absolute pixel-wise difference between adjacent frames was calculated. This allows the separation of data based on the intensity transitions between different patterns. The total number of frames for the checkerboard and star pattern was set to far more than 8 frames, and 8 middle frames were selected and averaged to ensure fairness evaluation against noise. 
%For rotating-star and phase pattern, the middle frame was selected for calibration. 

Calibration images was taken by the Intel RealSense D435 Infrared camera with three exposure levels($2000$, $4000$ and $8000\mu s$). In this experiment, rather than focusing on the reprojection error within the calibration dataset, we are more interested in evaluating performance under independent test views. To this end, we collected two datasets with different initial poses. Each dataset was captured by rotating the iPad around a fixed axis at 10\textdegree~step starting from its initial pose(see top row of Fig.~\ref{fig:ext_normal}). The rotation angle was controlled by a HUIKE mechanical displacement platform. We independently calibrate the camera using each dataset and then test on the other by solving for the pose using PnP algorithm and calculating the mean reprojection error(MRE). Notably, the second dataset contains more feature points that are distributed farther from the image center. Therefore, we expect that calibrating with the first dataset and testing on the second will pose a greater challenge for feature refinement algorithms.

\begin{figure*}[h!]
	\begin{center}
		\includegraphics[width=0.99\linewidth]{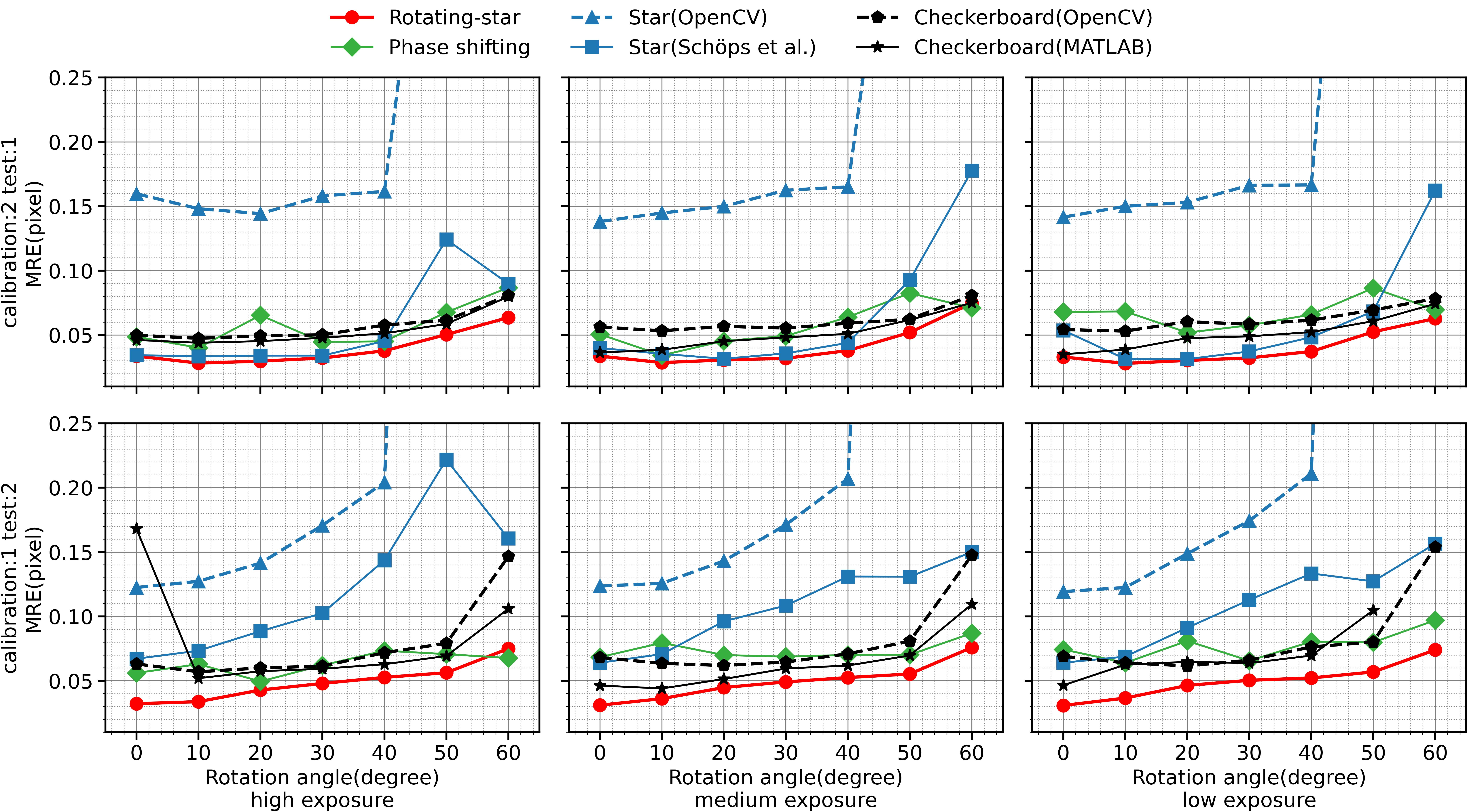}
	\end{center}
	\caption{Mean reprojection error evaluated using one group of poses for calibration and another group for testing. A half-window size of 10 pixels was used for optimization in all methods except for the phase-shifting approach.}
	\label{fig:normal_result}
\end{figure*}

As shown in bottom row of Fig.~\ref{fig:ext_normal}, all patterns from 0\textdegree~to 30\textdegree~exhibit a certain amount of overexposure. However, this does not significantly affect the calibration results of the phase-shifting method, see Fig.~\ref{fig:normal_result}. In fact, the results are better than those under the lowest exposure. This can be understood as a result of the higher contrast and lower noise brought by high exposure, which is essential for phase-shifting and consequently improves the overall MRE. To simplify the pipeline, the initial feature positions are provided by our method, whose reprojection error is generally above 0.1 pixels, thus the impact of initialization can be ignored. The most exposure-sensitive methods are the method of Sch\"ops~\etal~\cite{Schops2020} for star pattern and MATLAB for checkerboard, the latter shows abnormal reprojection errors at 0\textdegree~with high exposure in the second group, and fails to fully detect the checkerboard at 60\textdegree~under low exposure.

For the checkerboard, we compared the refinement results of OpenCV and MATLAB. As mentioned in the Sec.~\ref{sec:related_works}, the latter includes a preprocessing optimization for the saddle surface fitting. For the phase pattern, we adopted the optimization technique which fits a phase plane to neighboring pixels~\cite{Phase1}. However, this method of constraining the phase within a certain interval did not yield satisfactory results in our experiments. The best result, as shown in the Fig.~\ref{fig:normal_result}, was achieved using a fixed neighborhood with a half-window size of 3 pixels. 

As shown in Fig.~\ref{fig:normal_result}, our method achieved the best results at almost all tested positions, especially when the second group is used for test, where we applied the proposed refinement method(Eq.~\eqref{eq:cost}) here. The star-shaped pattern yielded results close to ours when the tilt angle in the first calibration group was below 50\textdegree, and it clearly outperformed the checkerboard method. However, its performance degraded significantly at larger angles and on the second calibration group, which involves larger rotational angles along the z-axis(see Fig.~\ref{fig:ext_normal}). This might be caused by more severe compression of the aliased region. As shown in the bottom row of Fig.~\ref{fig:ext_normal}, the aliased region of star pattern is visually elliptical, and becomes flattened as the tilt increases. Distortion can also affect this method, as mentioned by Sch\"ops~\etal in the conclusion of their paper~\cite{Schops2020}.

\section{Conclusion}
In this paper, we propose a novel active pattern for camera calibration along with a feature extraction pipeline. In the synthetic experiment, we demonstrate the impact of the PSF symmetry on the stability of refinement algorithms for checkerboard and rotating-star pattern, which indicates the possible improvement of proposed pattern for camera calibration with large aberration. In the real-world experiment, proposed method shows the accuracy advantages over the single star-shaped pattern and the phase-shifted method, especially under varying exposure settings and large calibration board tilt angles.

\clearpage
\bibliography{egbib}

\end{document}